\documentclass[10pt,twocolumn,letterpaper]{article}

\usepackage[pagenumbers]{cvpr} 

\usepackage{graphicx}
\usepackage{amsmath}
\usepackage{amssymb}
\usepackage{booktabs}

\usepackage{makecell, multirow}   
\usepackage{pifont}     
\usepackage{xcolor}     
\usepackage{pbox}       
\usepackage{booktabs}   

\usepackage[pagebackref,breaklinks,colorlinks]{hyperref}

\usepackage[capitalize]{cleveref}
\crefname{section}{Sec.}{Secs.}
\Crefname{section}{Section}{Sections}
\Crefname{table}{Table}{Tables}
\crefname{table}{Tab.}{Tabs.}

\def\cvprPaperID{15} 
\def\confName{CVPR}
\def\confYear{2023}

\newcommand{\ourschedule}{sketch plug-and-drop\xspace}

\begin{document}

\title{Reference-based Image Composition with Sketch \\ via Structure-aware Diffusion Model}


\author{
Kangyeol Kim\textsuperscript{\rm 1,2}\quad\quad 
Sunghyun Park\textsuperscript{\rm 1}\quad\quad
Junsoo Lee\textsuperscript{\rm 3}\quad\quad
Jaegul Choo\textsuperscript{\rm 1}\\ \\
\textsuperscript{\rm 1} KAIST\quad\quad
\textsuperscript{\rm 2} Letsur Inc.\quad\quad
\textsuperscript{\rm 3} Naver Webtoon\quad\quad \\
{\tt\small \{kangyeolk, psh01087, jchoo\}@kaist.ac.kr,} \quad\quad
{\tt\small junsoolee93@webtoonscorp.com}
}

\twocolumn[{
\renewcommand\twocolumn[1][]{#1}
\maketitle
\vspace{-1.0cm}
\begin{center}
    \centering
    \includegraphics[width=1.0\linewidth]{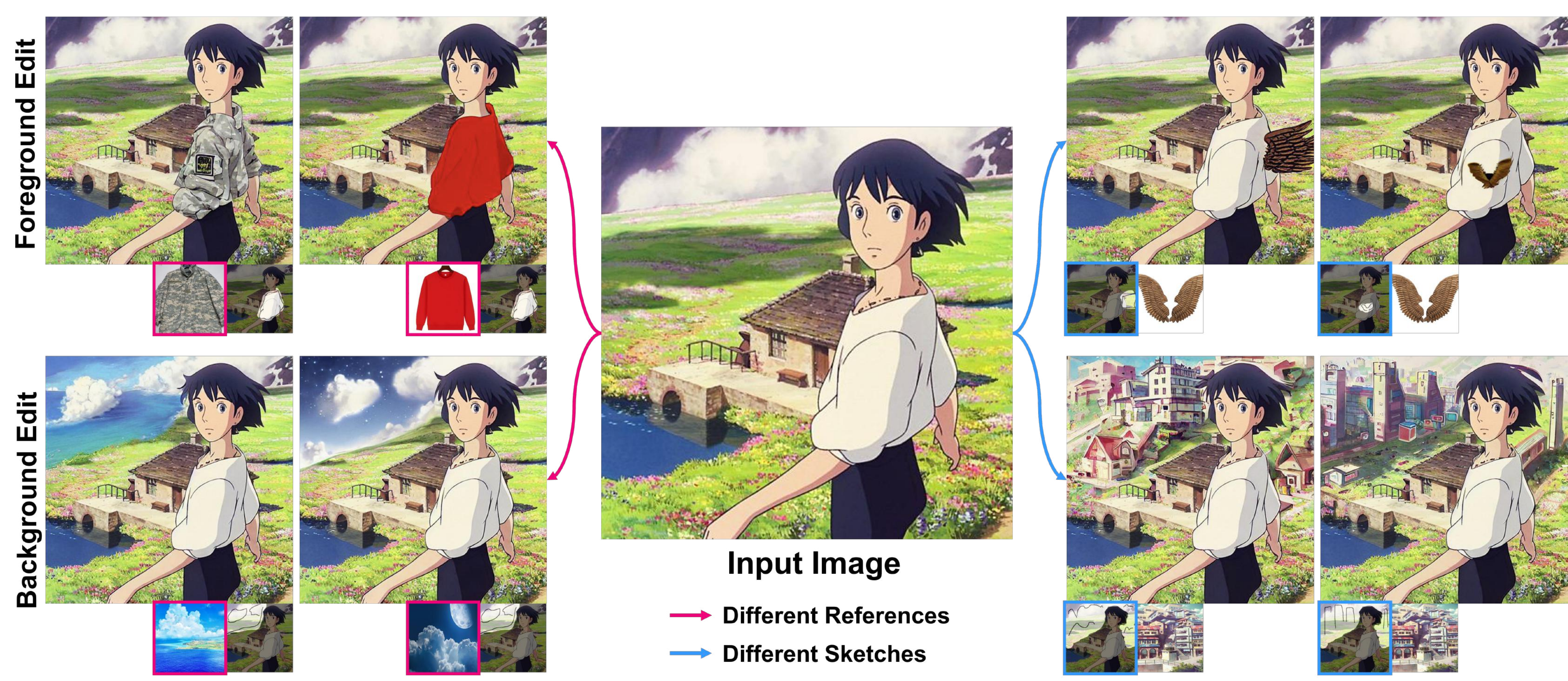}
    \vspace{-0.5cm}
    \captionof{figure}{Results of structure-aware diffusion model. Users are capable of editing  a scene by offering a reference image and sketch. 
    Our model can alter the foreground and background with a guidance of both modalities, and achieves a high-quality result.}
    \label{fig:teaser}
\end{center}
}]

\begin{abstract}
Recent remarkable improvements in large-scale text-to-image generative models have shown promising results in generating high-fidelity images.
To further enhance editability and enable fine-grained generation, we introduce a multi-input-conditioned image composition model that incorporates a sketch as a novel modal, alongside a reference image.
Thanks to the edge-level controllability using sketches, our method enables a user to edit or complete an image sub-part with a desired structure (\textit{i.e.,} sketch) and content (\textit{i.e.,} reference image).
Our framework fine-tunes a pre-trained diffusion model to complete missing regions using the reference image while maintaining sketch guidance.
Albeit simple, this leads to wide opportunities to fulfill user needs for obtaining the in-demand images.
Through extensive experiments, we demonstrate that our proposed method offers unique use cases for image manipulation, enabling user-driven modifications of arbitrary scenes.
\end{abstract}


\section{Introduction}

Recent advancements in large-scale text-to-image studies employing diffusion models~\cite{rombach2022ldm,ramesh2022dalle2,saharia2022imagen} have shown remarkable generative capabilities in synthesizing intricate images guided by textual input.
Building upon these foundational generative models, various approaches have been developed to enhance editability, either through the modification of a forward scheme during the inference process~\cite{meng2021sdedit} or by incorporating diverse modalities~\cite{zhang2023adding,yang2022paintbyexample,mou2023t2i}.
Notably, Paint-by-Example~\cite{yang2022paintbyexample} proposes the utilization of a visual hint to mitigate the ambiguity arising from textual descriptions. 
This empowers users to manipulate object-level semantics leveraging a reference image.

Our goal is to advance generative diffusion models by incorporating a partial sketch as a novel modal.
Sketches have long served as an intuitive and efficient means for creating a user-intended drawings, and are widely employed by both artists and the general population.
A key advantage of sketches compared to other modals, such as text~\cite{rombach2022ldm,ramesh2022dalle2,saharia2022imagen} and image~\cite{yang2022paintbyexample}, is to provide edge-level controllability by guiding the geometric structure during image synthesis.
This feature enables users to achieve finer detailed generation and editing of images in comparison to textual descriptions and standalone visual hints.
In practice, due to the significant utility of sketches in creating content in cartoons, this work focuses on the editing of cartoon scenes.


In this work, we propose a multi-input-conditioned image composition framework capable of generating a result guided by a sketch and reference image.
During generation, the sketch serves as a structure prior that determines the shape of the result within the target region.
To achieve this, we train a diffusion model~\cite{rombach2022ldm} to learn the completion of missing regions using the reference image, while maintaining sketch guidance.
Furthermore, we suggest a \textit{\ourschedule strategy} during the inference phase, which grants the model a degree of flexibility to relax sketch guidance.
The motivation behind this approach is to diminish the impact of overly simplified sketches (\textit{e.g.}, a single straight line for generating the clouds), and to make the model to accommodate a wide range of sketch types.

Compared to existing frameworks, sketch-guided generation offers distinguishable use cases for image manipulation.
Fig.~\ref{fig:teaser} presents visual examples utilizing distinct reference images and sketches.
In each row, the foreground and background have been modified separately by incorporating the provided conditions to fill in the target region.
These examples highlight the effectiveness of the proposed method in enabling user-driven modifications of arbitrary scenes.



\section{Methods}

\subsection{Preliminaries}

\noindent\textbf{Latent Diffusion Model.}
Recent text-to-image diffusion models such as LDM~\cite{rombach2022ldm} apply a diffusion model training in the latent space of a pre-trained autoencoder for efficient text-to-image generation.
Specifically, an encoder $\mathcal{E}$ encodes $\mathbf{x}$ into a latent representation $z=\mathcal{E}(\mathbf{x})$, and a decoder $\mathcal{D}$ reconstructs the image from $z$. 
Here, $\mathbf{x} \in \mathbb{R}^{3 \times H \times W}$ indicates an input image, where $H$ and $W$ denotes height and width, respectively.
Then, a conditional diffusion model $\epsilon_{\theta}$ is trained with the following loss function:
\begin{equation}
    \mathcal{L}=\mathbb{E}_{\epsilon(\mathbf{x}), \mathbf{y}, \epsilon \sim \mathcal{N}(0,1),t} \left[ \Vert \epsilon - \epsilon_{\theta} (z_t, t, \text{CLIP}(\mathbf{y})) \Vert^2_2 \right],
\end{equation}
where $\mathbf{y}$ denotes a text condition that is fed to a CLIP~\cite{radford2021clip} text encoder.
$t$ is uniformly sampled from $\left\{ 1, ..., T \right\}$, and $z_t$ is a noisy version of the latent representation $z$.
Moreover, a latent diffusion model employs a time-conditioned U-Net as $\epsilon_\theta$.
To achieve the faster convergence of our method, we employ Stable Diffusion~\cite{rombach2022ldm} as a strong prior.


\subsection{Proposed Approach}

\subsubsection{Training Phase}

\noindent\textbf{Problem Setup.}
We aim to train a diffusion model that takes the following inputs.
$\mathbf{x_p} \in \mathbb{R}^{3 \times H \times W}$ indicates an initial image, where $H$ and $W$ denotes height and width, respectively.
Let $\mathbf{m} \in \{0,1\}^{H \times W}$ denote a binary mask, where one indicates target editing regions, while zero means the regions to be preserved.
Corresponding sketch image $\mathbf{s} \in \{0,1\}^{H \times W}$ convey a structure information of masked region and a reference image $\mathbf{x_r} \in \mathbb{R}^{3 \times H' \times W'}$ is responsible for semantics inside the sketch.
During training, the model fills the masked regions following the sketch-guided structure with the contents of the reference image.

\noindent\textbf{Initialization.} 
During training, the model is responsible for generating the masked region following the sketch guidance and properly placing the reference image.
It may be extra tasks for model, instead we opt to use previous work~\cite{yang2022paintbyexample}'s trained weights as an initialization.
By doing this, the model has a strong prior to bring the reference image on the masked region.
The initialization makes the model achieve its objective at ease, by optimizing the initialized weights to follow the sketch guidance.
We found that the model takes a longer time to converge without the strong prior.


\noindent\textbf{Model Forward.}
We take self-supervised training to train a diffusion model.
For each iteration, the training batch consists of $\{\mathbf{x_p}, \mathbf{m}, \mathbf{s}, \mathbf{x_r}\}$, and the goal of the model is to properly produce the masked part $\mathbf{m} \odot \mathbf{x_p}$.
We \textit{randomly} generate a region of interest (RoI) as a bounding box, in which mask shape augmentation as previous work~\cite{yang2022paintbyexample} is applied to simulate a drawn-like mask.
On another branch, $\mathbf{x_r}$ is generated by cropping and augmenting the RoI region, being successively fed to a CLIP~\cite{radford2021clip} image encoder to make a condition $\mathbf{c}$ for a diffusion model.
Formally, it can be written as $\mathbf{c} = \text{MLP}(\text{CLIP}(\mathbf{x_r}))$, where MLP is a simple feed-forward network to transform the output distribution to be properly adjusted as the condition of the diffusion model.
For each diffusion step, the masked initial image, the sketch, and the previous step's result $\mathbf{y_t}$ are concatenated and fed into the diffusion model.

\subsubsection{Inference Phase}

\noindent\textbf{Sketch Plug-and-Drop Strategy.}
Although the free-drawn sketch is a handy condition for a user, the model occasionally has difficulty in strictly keeping the outline structure.
This is noticeable when it comes to generating scenery backgrounds such as clouds and snowy trees, where the boundaries are rather ambiguous.
In these cases, a simple straight line may be inadequate though the user's burden can be minimized.
In this respect, we add on a simple yet effective method, \textit{sketch \ourschedule}, in which the infusion steps of the sketch condition are flexibly adjusted.

\noindent\textbf{Self-reference Generation.}
Sketch-guided generation can be used in various cases, such as manipulating the shapes of objects and changing the poses.
When generating specific parts of an object, obtaining a suitable reference image is not trivial, because it is difficult to collect a harmonic image with a masked part.
In practice, we found that using a certain part of the initial image alternatively is a reasonable way to get the reference image.

\begin{figure}[t!]
    \centering
    \includegraphics[width=1.0\linewidth]{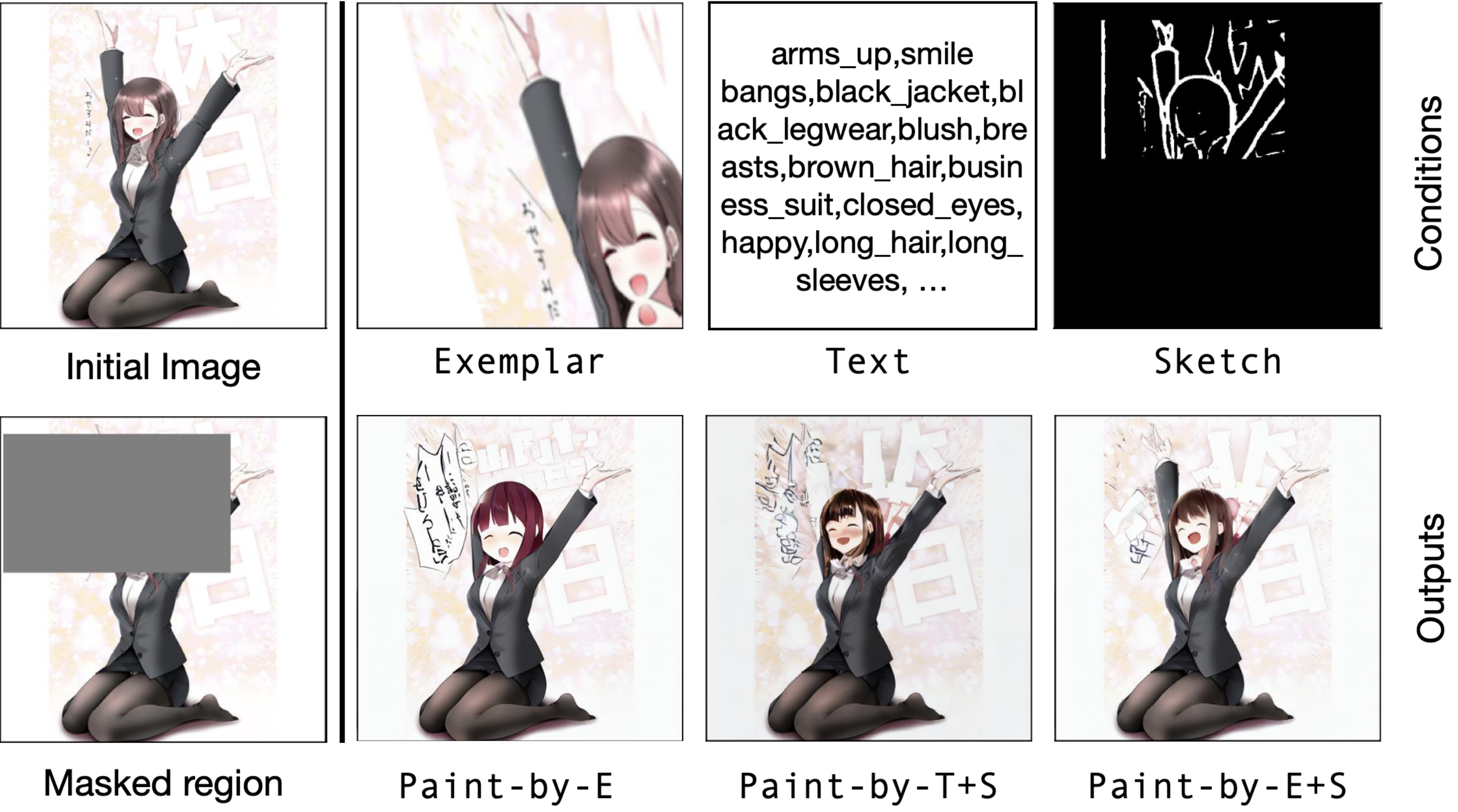}
    \vspace{-0.7cm}
    \caption{Qualitative comparisons with the baselines.}    
    \vspace{-0.3cm}
    \label{fig:qualitative-baselines}
\end{figure}

\begin{table}[t!]    
    \centering
    \begin{tabular}{@{} p{0.15\textwidth}<{\centering} p{0.075\textwidth}<{\centering} p{0.075\textwidth}<{\centering} p{0.075\textwidth}<{\centering}}
    \toprule[1.5pt]
     & ${L_1 \ error}$ & ${L_2 \ error}$ & ${FID}$ \\
    \midrule
    $\texttt{Paint-by-E}$ & 0.0866 & 0.0380 & 6.314 \\
    $\texttt{Paint-by-T+S}$ & 0.0851 & 0.0313 & 6.314 \\
    $\texttt{Paint-by-E+S}$ & \textbf{0.0680} & \textbf{0.2393} & \textbf{5.716} \\
    \bottomrule[1.5pt]
    \end{tabular}
    \vspace{-0.3cm}
    \caption{Quantitative comparisons with the baselines}
    \vspace{-0.5cm}
    \label{Table:comparison_baselines}
\end{table}

\section{Experiments}

As a training and testing dataset, we utilized Danbooru~\cite{danbooru2021} dataset.
Due to the massive volume of the original dataset, we opt to use its subset to reduce the excessive training duration. 
The Danbooru dataset encompasses a wide variety of animated characters, exhibiting diverse artistic styles from numerous artists. 
We employed a recently released edge detection method~\cite{su2021pidinet} to extract the edges, subsequently binarizing the extracted edges. 
The training and testing datasets comprise 55,104 and 13,775 image-sketch pairs, respectively.
For qualitative evaluation, we collect real-world cartoon scenes to showcase the potential of our work.
The majority of these cartoon scenes were sourced from Naver Webtoon platform~\footnote{https://comic.naver.com/webtoon/weekday} and captured from Ghibli studio's movies~\footnote{https://www.ghibli.jp/ \vspace{-1.1cm}}.

\subsection{Comparisons with Baselines}

\noindent\textbf{Baselines.}
To the best of our understanding, no prior research has proposed a multi-input-conditioned model with a diffusion model approach.
Therefore, we implement two baselines to analyze our model in a qualitative and quantitative manner.
In specific, we implement (1) \texttt{Paint-by-T+S} that uses a text-sketch pair instead of an example-sketch pair (2) \texttt{Paint-by-E(xample)}~\cite{yang2022paintbyexample} to reveal the effectiveness of sketch guidance to complete a missing part of an image.
One of our interests is to demonstrate the superiority of an example-sketch pair compared to other guidance.
In the following experiments, we focus on unraveling the potential of such guidance by not only showing superb quantitative results but also providing multiple use cases of our model.
All baselines and our model are trained with the same configuration.

For quantitative comparison, we use the averages of $L_1$ and $L_2$ errors between the initial and reconstructed images.
We utilize Frechét inception distance (FID)~\cite{heusel2017gans} to evaluate the visual quality of the generated images.

\begin{figure}[t!]
    \centering
    \includegraphics[width=1.0\linewidth]{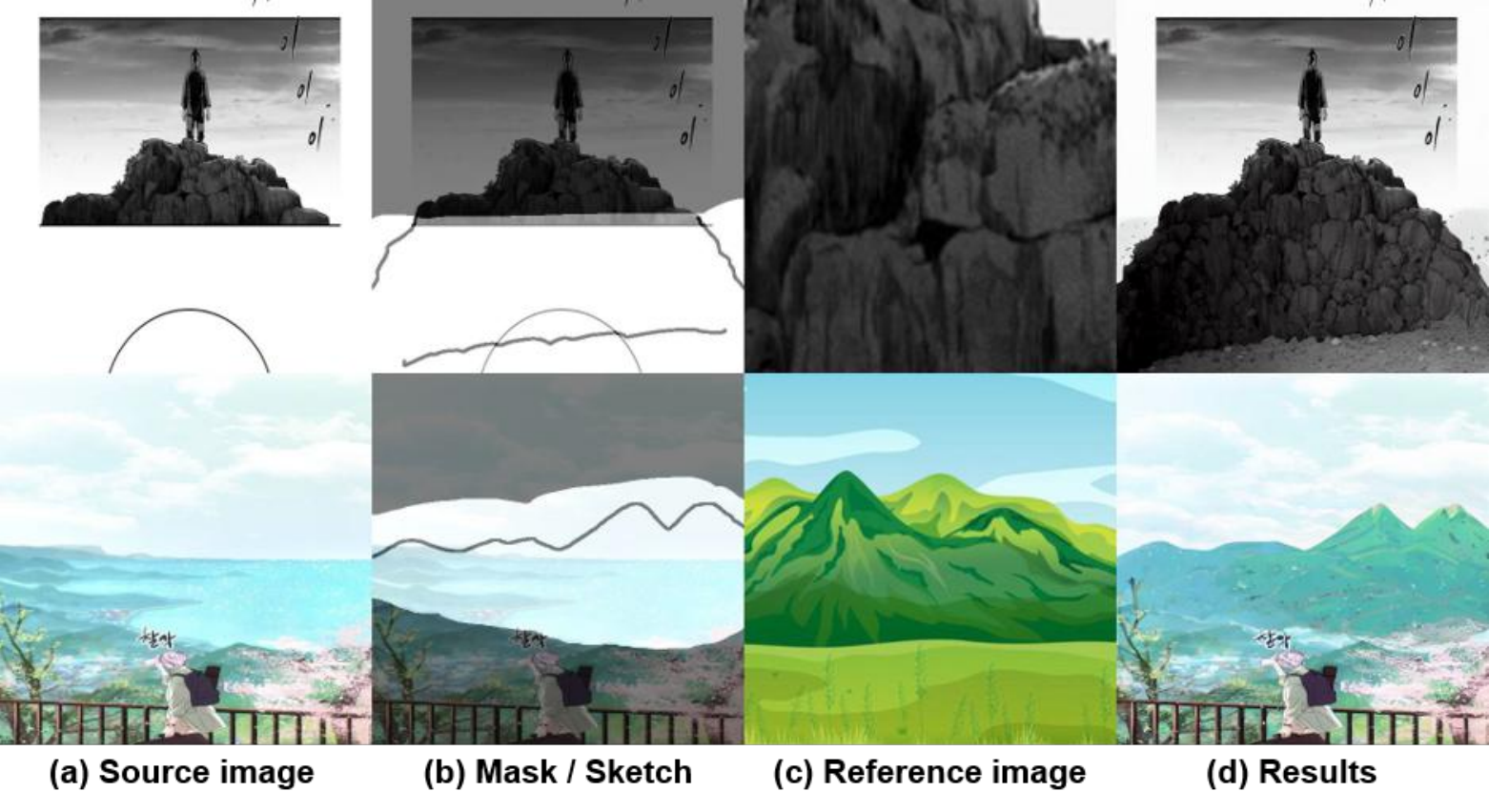}
    \vspace{-0.75cm}
    \caption{Examples for background scene editing.}    
    \vspace{-0.2cm}
    \label{fig:scene-edit}
\end{figure}

\begin{figure}[t!]
    \centering
    \includegraphics[width=1.0\linewidth]{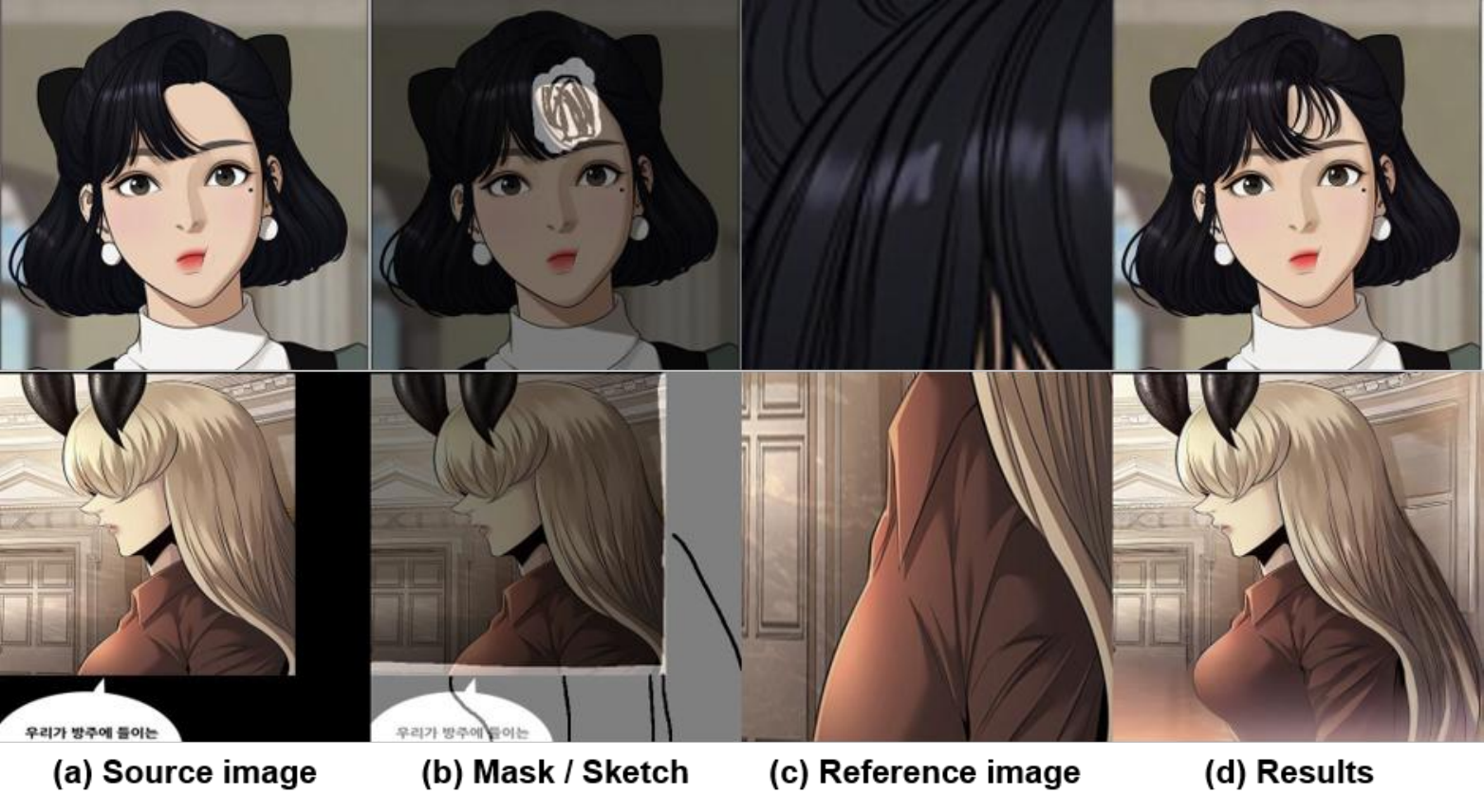}
    \vspace{-0.75cm}
    \caption{Examples of local object shape editing applications.}    
    \vspace{-0.5cm}
    \label{fig:local-edit}
\end{figure}

\noindent\textbf{Comparison Results.}
Fig.~\ref{fig:qualitative-baselines} shows obvious differences in each input setting.
Using a sole reference image is insufficient to make a good guess of the missing part, producing an aesthetically unappealing completion result (\textit{2nd} column of Fig.~\ref{fig:qualitative-baselines}).
On the other hand, simply feeding sketch input greatly improves visual quality by guiding the structure. 
Especially, unlike a text condition that generally contains information for the entire image, an exemplar image could be an efficient condition for filling the local context.

Table~\ref{Table:comparison_baselines} shows quantitative comparisons with baselines.
As can be seen, \texttt{Paint-by-E} relatively performs worse than other models, because there is no explicit guidance about the structure within the masked region.
Compared to it, both \texttt{Paint-by-T+S} and \texttt{Paint-by-E+S} exhibit superior performances thanks to the sketch conditions.
Particularly, \texttt{Paint-by-E+S} approach demonstrates the most exceptional performance, in conjunction with accompanying sketch and exemplar image.

\begin{figure}[t!]
    \centering
    \includegraphics[width=1.0\linewidth]{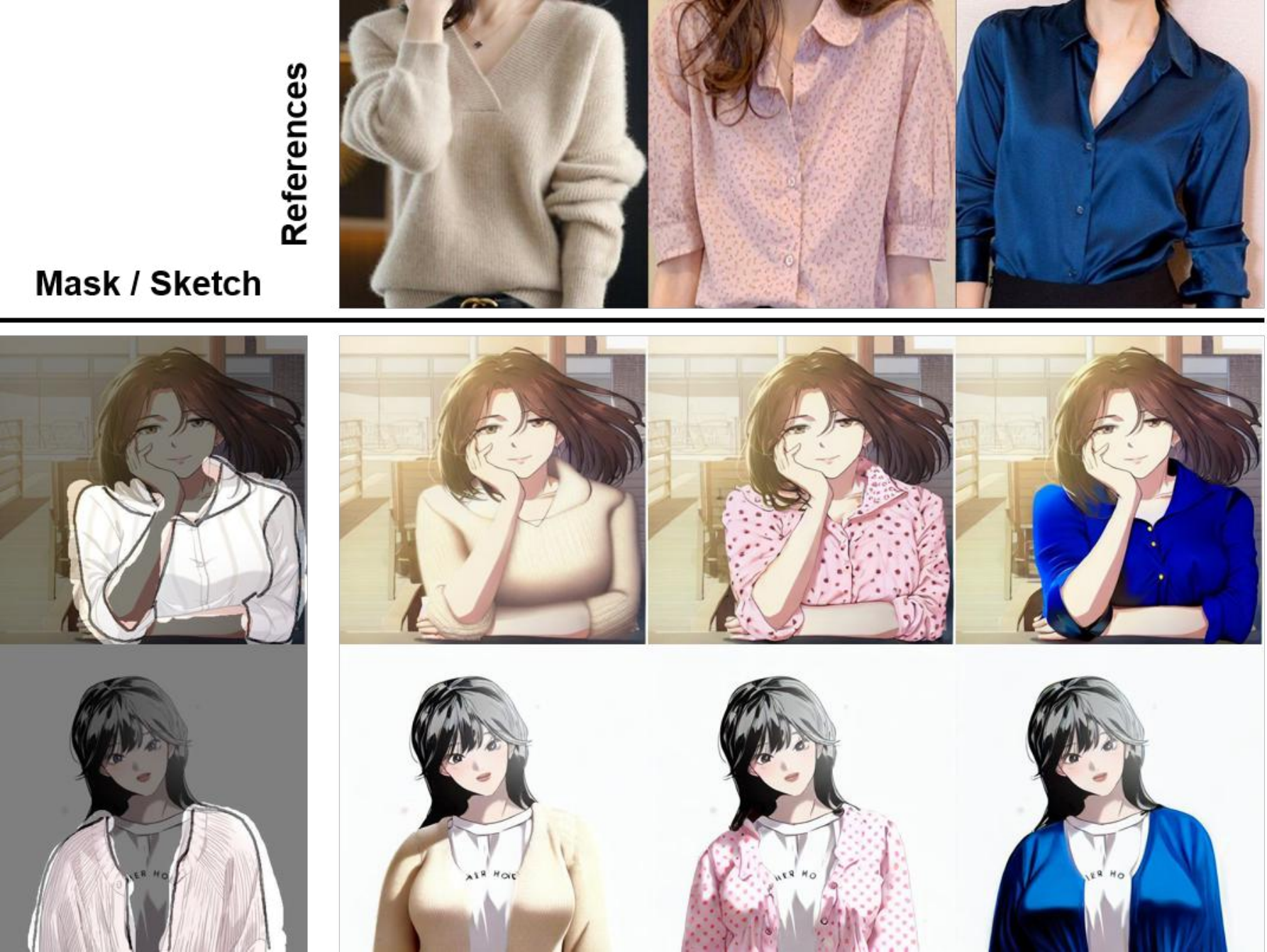}
    \vspace{-0.7cm}
    \caption{Visual results of object change with reference images.}
    \vspace{-0.2cm}
    \label{fig:multi-references}
\end{figure}

\begin{figure}[t!]
    \centering
    \includegraphics[width=1.0\linewidth]{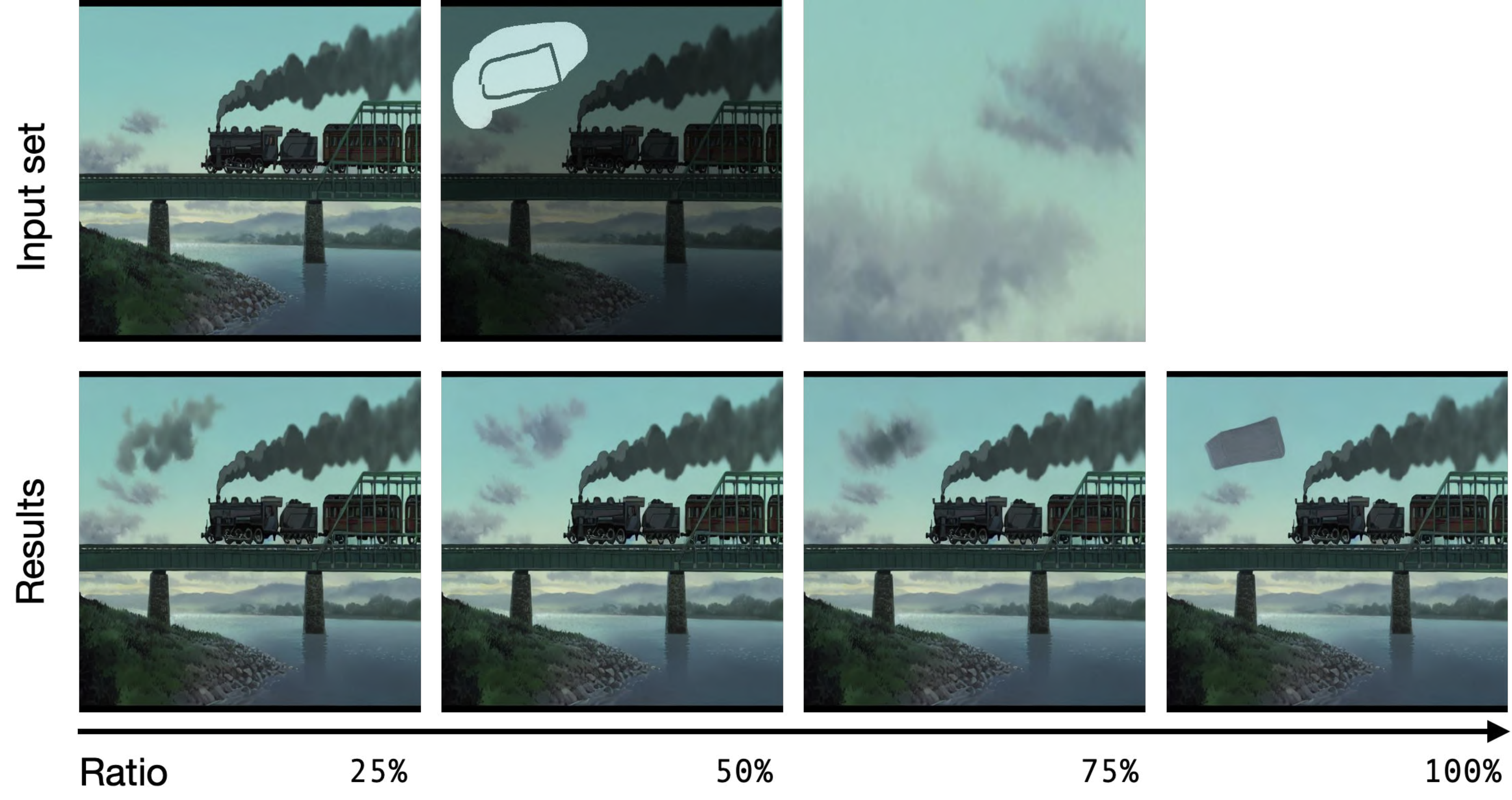}
    \vspace{-0.7cm}
    \caption{Effect of sketch \ourschedule strategy.}
    \vspace{-0.5cm}
    \label{fig:sketch-timestep-control}
\end{figure}

\subsection{Application Scenarios}
In this section, we show multiple representative applications of our model for editing real-world cartoon scenes.
Note that since a sketch has a variety of shapes, the applications are not limited to presenting examples.

\noindent\textbf{Background Scene Editing.}
Drawing background cartoon scenes is labor-intensive and time-consuming work.
Hence, many scenes are cropped on purpose and reduce the author's effort to draw scenery parts.
In response, our approach opens the way to complete and extend the cropped scenes, giving a chance to flexibly control a shape and semantics.
Fig~\ref{fig:scene-edit} shows that new continuous scenes have been successfully added to real-world cartoon ones.
This enables the authors not to be dedicated to creating unimportant scenes.

\noindent\textbf{Object Shape Editing.}
Fig.~\ref{fig:local-edit} shows that our model's use case is to edit fine-detailed object shapes such as hairs and beards.
As can be seen, a user can manipulate the structure of local regions by simply giving user-desirable sketches.
This application is practically useful for generating numerous scenes that have different structures.

\noindent\textbf{Object Changes.}
Our model takes a reference image that is used to determine the in-context of a masked region.
In this sense, a preferable reference image from a user serves to generate user-desirable images.
As shown in Fig.~\ref{fig:multi-references}, we can readily alter an upper cloth of a character by providing various references.
Surprisingly, a texture or pattern of cloth is imported to the generated results as well as reference colors.

\begin{figure}[t!]
    \centering
    \includegraphics[width=1.0\linewidth]{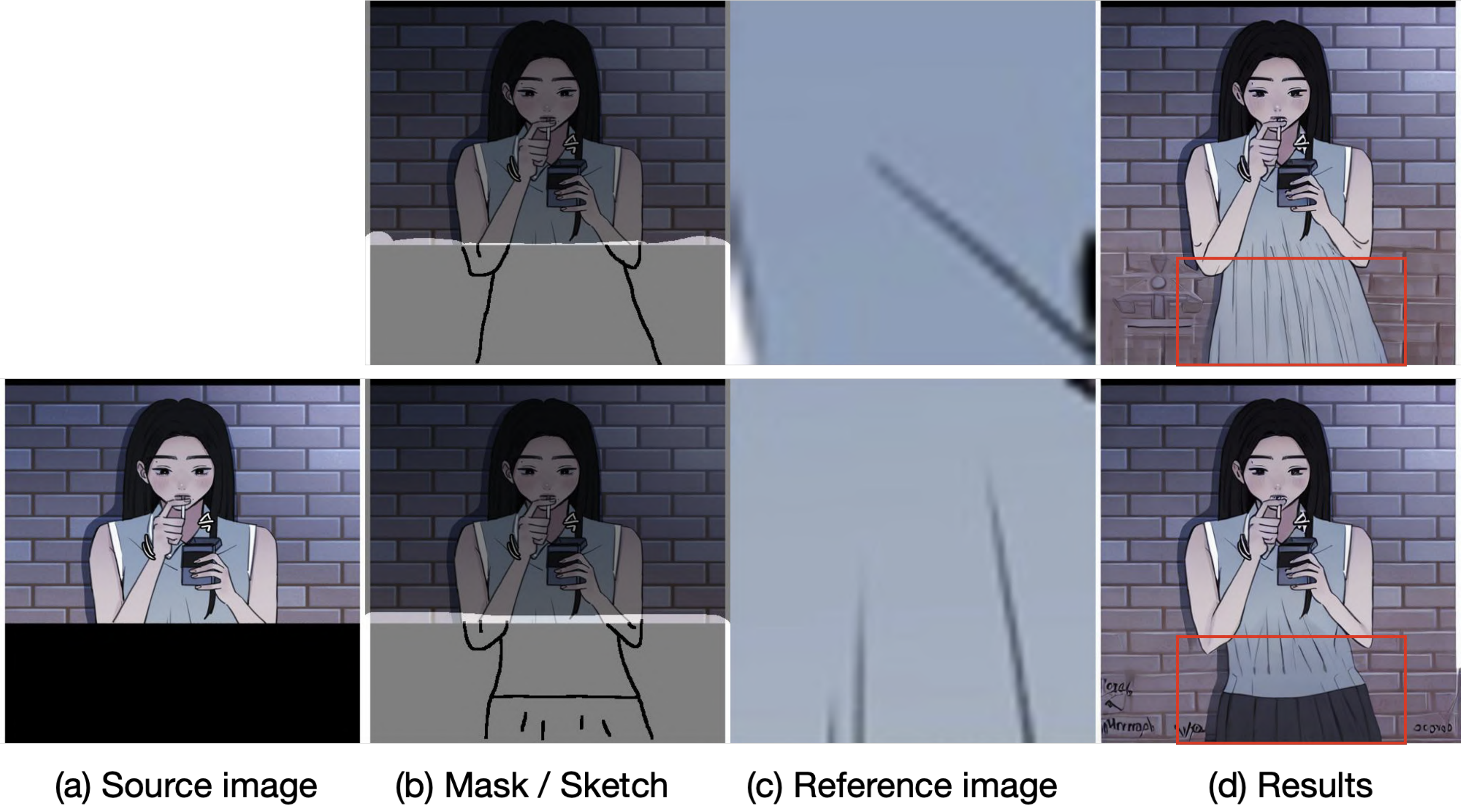}
    \vspace{-0.7cm}
    \caption{Effect of different sketches for object synthesis.}
    \vspace{-0.5cm}
    \label{fig:sketch-control}
\end{figure}

\subsection{Qualitative Analysis}

\noindent\textbf{Effect of Sketch Plug-and-Drop.}
Fig.~\ref{fig:sketch-timestep-control} presents the effect of the \ourschedule strategy.
In this case, a user means to add a cloud to the sky, yet the sketch guidance is composed of straight lines that are not suitable to represent the detailed boundaries of a cloud.
As a result, the synthesized cloud is awkward as seen in the last column of Fig~\ref{fig:sketch-timestep-control}.
On the other hand, reducing the time-step is an effective workaround to relax an over-constrained sketch condition, leading to more natural results as presented in the rest columns of Fig~\ref{fig:sketch-timestep-control}.

\noindent\textbf{Effect of Sketch Boundary.}
A sketch with multiple lines forms boundaries in an image, and we found that the boundaries act as a pivotal point.
As shown in Fig.~\ref{fig:sketch-control}, two sketches bring on different results, especially a straight line of the second sketch serves to determine the boundary of cloths.

\section{Discussions and Conclusions}
In this paper, we present a novel sketch-guided diffusion model.
Our primary motivation lies in fully utilizing \textit{partial} sketch and reference image during diffusion process to enable a user to control the structure of output.
With our model, a user successfully generates and manipulate the targeted region, conditioned on a user-drawn sketch and reference image.
Throughout the generation process, the sketch offers structural guidance, while the reference image dictates the output's semantics.
We demonstrated the utilities of the proposed approach by showing various use examples.
Despite its effectiveness, our model can be further enhanced to provide a more user-friendly tool in practical scenarios.
Given the consideration of multiple inputs, a user-centric system that facilitates seamless interaction between the user and the model needs to be explored 
In following research, we plan to address this issue, and devise a highly intuitive tool incorporating our model.

{\small
\bibliographystyle{ieee_fullname}
\bibliography{reference}
}


\twocolumn[
\begin{center}
    \vspace*{0.6cm}
    \Large{\bf{Supplementary Material}}
    \vspace*{1.7cm}
\end{center}]

\section*{A. Implementation Details}

\begin{figure}[t!]
    \centering
    \includegraphics[width=1.0\linewidth]{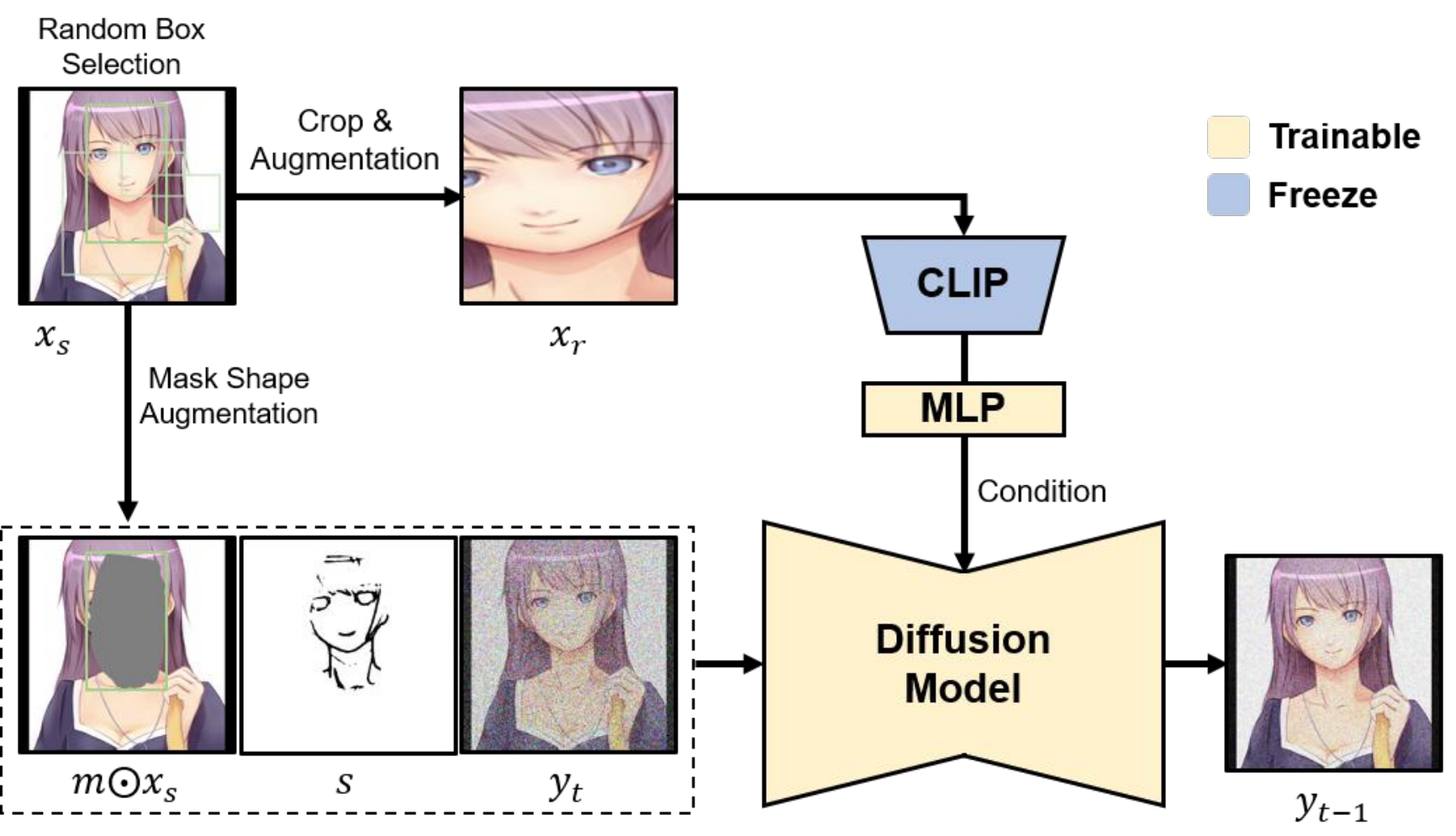}
    \vspace{-0.4cm}
    \caption{Model overview.}
    \vspace{-0.5cm}
    \label{fig:model overview}
\end{figure}

To better utilize prior knowledge and accelerate the training process, we adapt pre-trained weights of Paint-by-Example~\cite{yang2022paintbyexample} model as initialization.
Because Paint-by-Example~\cite{yang2022paintbyexample} utilized a well-known text-to-image generation model, Stable Diffusion~\cite{rombach2022ldm}, as initialization, our implementation also makes use of a strong prior knowledge of seminal work.
During the training process, the images are resized to a size of $512\times512$, and the model is trained for a total of 40 epochs, with a duration of approximately 2 days when using 4 NVIDIA V100 GPUs.
We utilized the AdamW~\cite{loshchilov2017decoupled} optimizer with a learning rate of 1e-5 and set the batch size to 4.
Fig.~\ref{fig:model overview} shows the overview of the training process of the proposed method.

\section*{B. Large-scale Diffusion Models}



After a latent diffusion model (LDM)~\cite{rombach2022ldm} shows remarkable performance on the unconditional generation task, many studies explore how to obtain the generative results that users want by preserving the knowledge of LDM as a prior.
To make the generation output contain a specific object, Dreambooth~\cite{ruiz2022dreambooth} and TextualInversion~\cite{gal2022textual-inversion} adopt the finetuning strategy; the former makes all parameters trainable, but the latter only finetunes a special token.
GLIGEN~\cite{li2023gligen} introduces a new gated layer to address new input modalities, such as bounding boxes and keypoints.
Plug-and-Play~\cite{tumanyan2022plugandplay} shows a promising result to control the shape and texture of the output without any finetuning or introducing new layers.
Recently, Paint-by-Example~\cite{yang2022paintbyexample} shows a way to replace a text condition with an image, which makes it easy to inject the object or even style information if its reference image exists.
The over-mentioned methods tackle controlling the structure of the target object; however, it cannot be guaranteed completely.
As a concurrent work, several studies such as ControlNet~\cite{zhang2023adding} and Composer~\cite{lhhuang2023composer} have demonstrated the ability to generate images from sketches. 
However, their primary emphasis lies in utilizing sketches as the guidance for the image creation process, rather than in editing images.
In contrast, our model takes a cartoon image as input and performs selective modifications on a specific region while preserving the integrity of the remaining regions of the image.

\section*{C. Additional Experiments}

\begin{figure}[h!]
    \centering
    \includegraphics[width=1.0\linewidth]{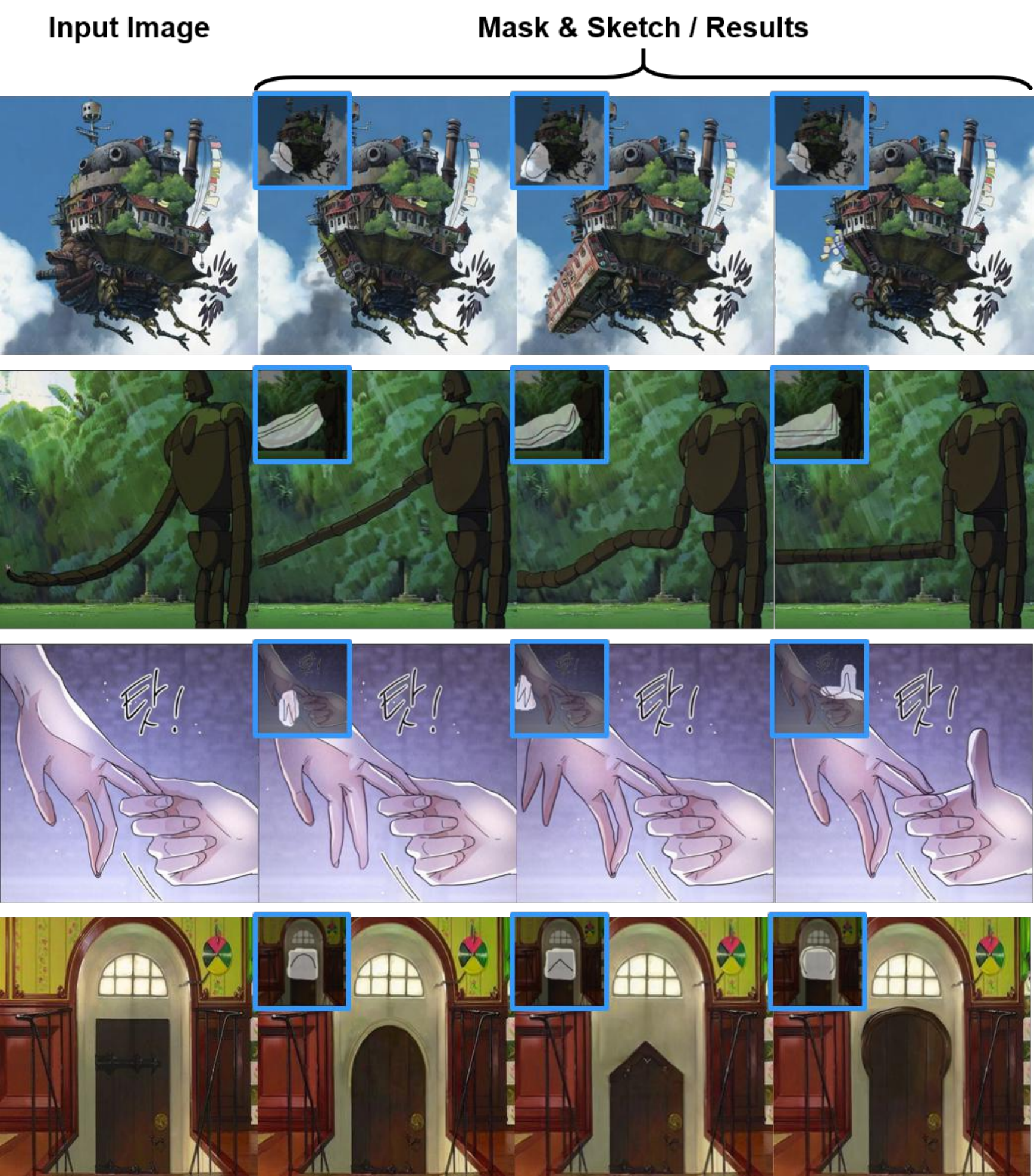}
    \vspace{-0.7cm}
    \caption{Additional results of local object shape editing with different sketches.}
    \vspace{-0.7cm}
    \label{supp-fig:local-edits}
\end{figure}

\begin{figure*}[t!]
    \centering
    \includegraphics[width=0.9\linewidth]{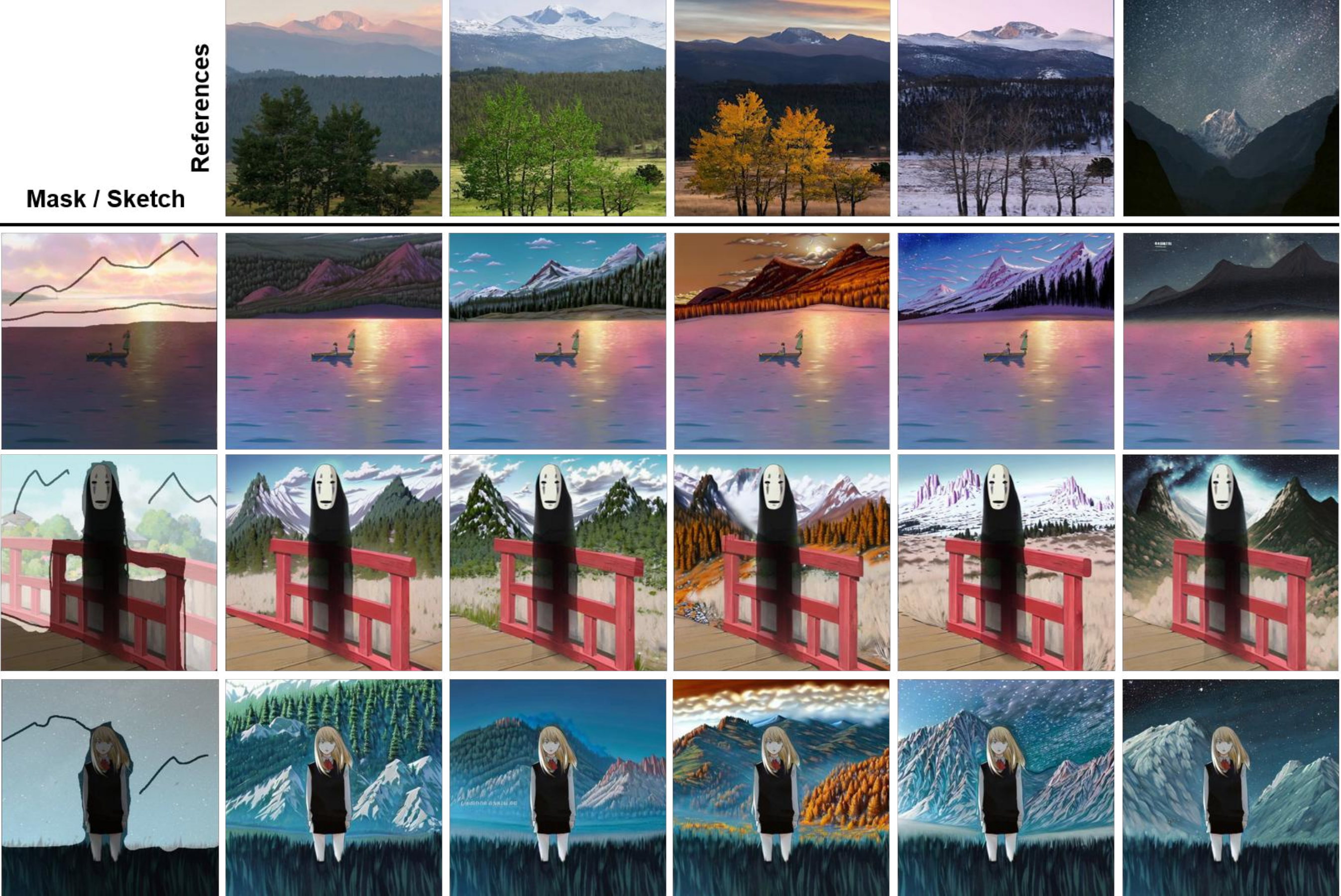}
    \vspace{-0.2cm}
    \caption{Additional results of background scene editing with reference images.}
    \vspace{-0.1cm}
    \label{supp-fig:multi-backgrounds}
\end{figure*}

\begin{figure*}[t!]
    \centering
    \includegraphics[width=0.9\linewidth]{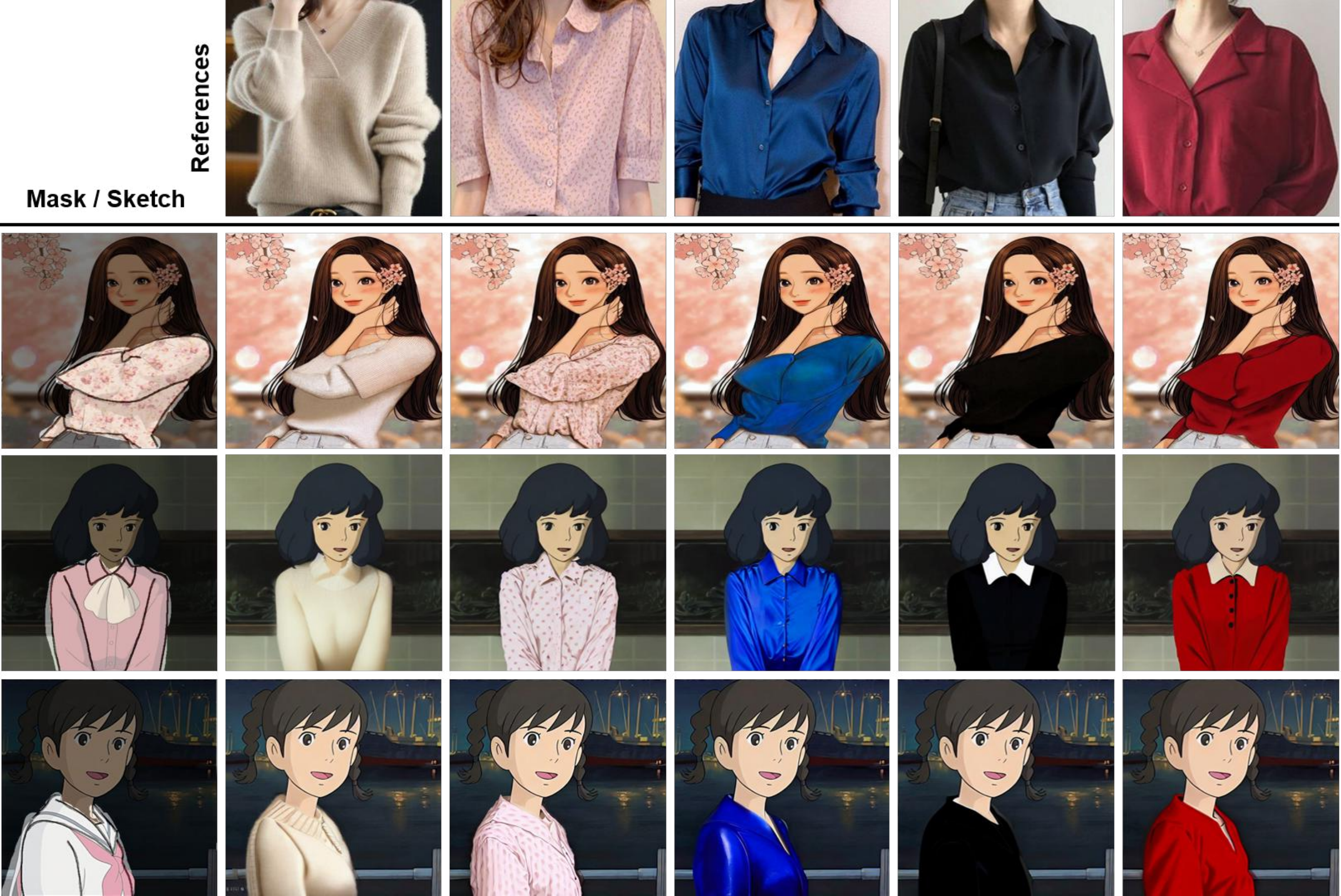}
    \vspace{-0.2cm}
    \caption{Additional results of object change with reference images.}
    \vspace{-0.1cm}
    \label{supp-fig:multi-references}
\end{figure*}

We provide additional results to demonstrate the applicability of our method for editing real-world cartoon scenes sourced from Naver webtoon and Studio Ghibli films.
Fig.~\ref{supp-fig:local-edits}, \ref{supp-fig:multi-backgrounds}, and \ref{supp-fig:multi-references} show the additional results of our method.

\end{document}